\documentclass[conference]{IEEEtran}
\IEEEoverridecommandlockouts
\usepackage{cite}
\usepackage{textcomp}
\usepackage[hyphens]{url}  
\usepackage{graphicx} 
\usepackage{subcaption}
\usepackage{caption} 
\usepackage{algorithm}
\usepackage{algorithmic}
\usepackage{enumitem}
\usepackage{amsmath}
\usepackage{array}
\usepackage{makecell}
\usepackage{multirow}
\usepackage{adjustbox}
\usepackage{xcolor}
\usepackage{newfloat}
\usepackage{listings}
\title{\textsc{\textit{MoENAS}}: Mixture-of-Expert based Neural Architecture Search for jointly Accurate, Fair, and Robust Edge Deep Neural Networks}

\author{%
  Lotfi Abdelkrim Mecharbat\textsuperscript{1}, Alberto Marchisio\textsuperscript{1,2}, Muhammad Shafique\textsuperscript{1,2}, Mohammad M. Ghassemi\textsuperscript{3}, \\
  Tuka Alhanai\textsuperscript{1} \\
  \textsuperscript{1}Center for Quantum and Topological Systems (CQTS), NYUAD Research Institute, NYUAD, Abu Dhabi, UAE\\
  \textsuperscript{2}eBRAIN Lab, Division of Engineering, New York University Abu Dhabi (NYUAD), Abu Dhabi, UAE\\
  \textsuperscript{3}Department of Computer Science, Michigan State University, MI USA\\
  \texttt{lam9297@nyu.edu, alberto.marchisio@nyu.edu, muhammad.shafique@nyu.edu,}\\
  \texttt{ghassem3@msu.edu, tuka.alhanai@nyu.edu} \\
}

\begin{document}

\maketitle

\begin{abstract}

There has been a surge in optimizing edge Deep Neural Networks (DNNs) for accuracy and efficiency using traditional optimization techniques such as pruning, and more recently, employing automatic design methodologies. However, the focus of these design techniques has often overlooked critical metrics such as fairness, robustness, and generalization. As a result, when evaluating SOTA edge DNNs' performance in image classification using the FACET dataset, we found that they exhibit significant accuracy disparities (14.09\%) across 10 different skin tones, alongside issues of non-robustness and poor generalizability. In response to these observations, we introduce \textit{Mixture-of-Experts-based Neural Architecture Search (MoENAS)}, an automatic design technique that navigates through a space of mixture of experts to discover accurate, fair, robust, and general edge DNNs. \textit{MoENAS} improves the accuracy by 4.02\% compared to SOTA edge DNNs and reduces the skin tone accuracy disparities from 14.09\% to 5.60\%, while enhancing robustness by 3.80\% and minimizing overfitting to 0.21\%, all while keeping model size close to state-of-the-art models average size (+0.4M). With these improvements, \textit{MoENAS} establishes a new benchmark for edge DNN design, paving the way for the development of more inclusive and robust edge DNNs.

\

\end{abstract}

\section{Introduction}
\label{sec:introduction}
In recent years, the growing use of deep learning across various fields~\cite{yang2021intelligent, macas2022survey, grigorescu2020survey} has highlighted the need for efficient, safe, and private deployment methods, driving the adoption of edge computing. By bringing computation closer to the data source, edge computing reduces latency, saves bandwidth, and enhances privacy and security. However, deploying deep neural networks (DNNs) on edge devices poses challenges due to limited computational resources and energy constraints~\cite{alvear2023edge}.

To tackle these challenges, researchers have focused on improving the accuracy and efficiency of DNNs for edge applications. Traditional approaches, such as manual model optimization~\cite{Marchisio_2018IJCNN_PruNet, vadera2021methods, Hanif_2022MICPRO_EfficientEmbeddedDL, chen2021quantization, matsubara2020head}, often fail to balance multiple objectives effectively. The emergence of Hardware-aware Neural Architecture Search (HW-NAS) has automated the search for optimal architectures, improving model efficiency and accuracy in resource-constrained settings~\cite{benmeziane2021comprehensive}.

HW-NAS uses machine learning to explore architecture spaces and identify designs that balance performance and resource consumption, accelerating edge DNN deployment~\cite{zhang2020fast}. Unlike traditional NAS, which prioritizes accuracy~\cite{he2021automl, elsken2019neural}, HW-NAS employs multi-objective optimization to enhance both accuracy and efficiency (e.g., latency, size)\cite{wistuba2019survey, Marchisio_2020ICCAD_NASCaps, Prabakaran_2021JIOT_BioNetExplorer}, generating Pareto-optimal solutions\cite{Kaisa1999MultiObjective}.

\subsection{Target Research Problem and Associated Challenges}

While advancements in edge DNN design have improved accuracy and computational efficiency, critical performance metrics like fairness, robustness, and generalization remain underexplored~\cite{sheng2022larger}. Fairness ensures equitable performance across diverse user groups, robustness measures reliability under varying conditions (e.g., lighting, weather, visibility)\cite{drenkow2021systematic, porrello2020robust}, and generalization evaluates performance on unseen data\cite{zhou2022domain}.

Enhancing these metrics is essential for edge DNNs, particularly in safety-critical applications like medical diagnostics~\cite{esteva2021deep}. However, achieving this faces challenges such as (1) ensuring data quality and balance~\cite{mehrabi2021survey}, (2) addressing insufficient data diversity~\cite{recht2019imagenet}, and (3) overcoming the computational limitations of edge devices~\cite{ibrahim2022robustness, feuerriegel2020fair, hickey2021fairness}.

Most existing work addresses these challenges by improving data quality~\cite{pitoura2020social} or training procedures~\cite{jain2024fairness}, with limited focus on architecture design. Studies targeting architecture often prioritize a single metric, such as fairness~\cite{sheng2022larger}. Comprehensive approaches that consider fairness, robustness, and generalization together are scarce. To fill this gap, we aim to propose a design methodology ensuring edge DNNs are not only efficient and high-performing but also fair and robust.

\subsection{Analysis: Fairness, Robustness, Generalization in Edge DNNs} \label{intro:pre-analysis}
To highlight the limitations of edge DNN designs in addressing fairness, robustness, and generalization, we evaluated 12 state-of-the-art (SOTA) edge DNNs on person classification (binary classification of images with or without a person) using a COCO dataset subset~\cite{lin2014microsoft}. We measured overfitting, sensitivity to light, and accuracy across different skin tones using the FACET dataset~\cite{gustafson2023facet}.

\begin{figure}
\centering
\begin{subfigure}[b]{0.48\linewidth}
\includegraphics[width=1.\linewidth]{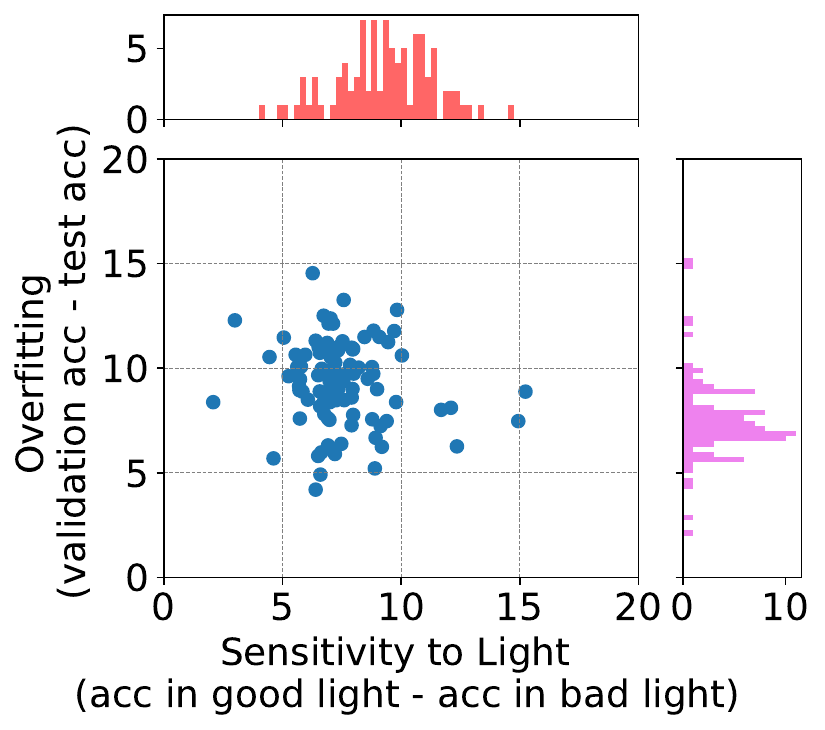}
    \caption{Overfitting and Sensitivity to light conditions of edge SOTA DNNs.}
    \label{fig:sota_rob}
    \end{subfigure}
\hfill
\begin{subfigure}[b]{0.48\linewidth}
\includegraphics[width=1.\linewidth]{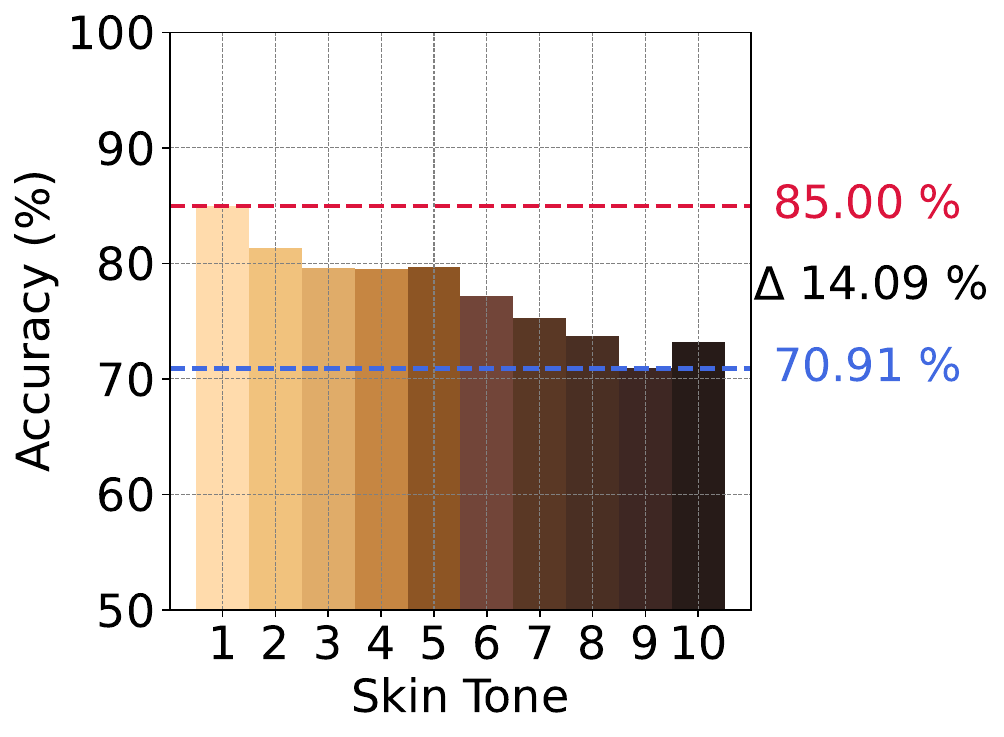}
    \caption{Average Accuracy of SOTA edge DNNs across 10 levels of skin tones (from lightest=1 to darkest=10) on FACET~\cite{gustafson2023facet}. }
    \label{fig:sota_fair}
\end{subfigure} 
    \caption{Evaluation of Fairness, Robustness, and Generalization of SOTA edge DNNs on FACET.}
    \label{fig:sota}
\end{figure}

Figure~\ref{fig:sota_rob} presents robustness and generalization results, revealing notable gaps: validation and test accuracies differ significantly, and accuracies under varying lighting conditions drop by up to 15\%. This indicates potential overfitting and poor robustness to light changes in SOTA models. Figure~\ref{fig:sota_fair} shows a concerning fairness issue: average accuracy declines from 85.0\% for the lightest skin tone category to 70.9\% for the darkest.

\begin{figure}
    \centering
    \includegraphics[width=\linewidth]{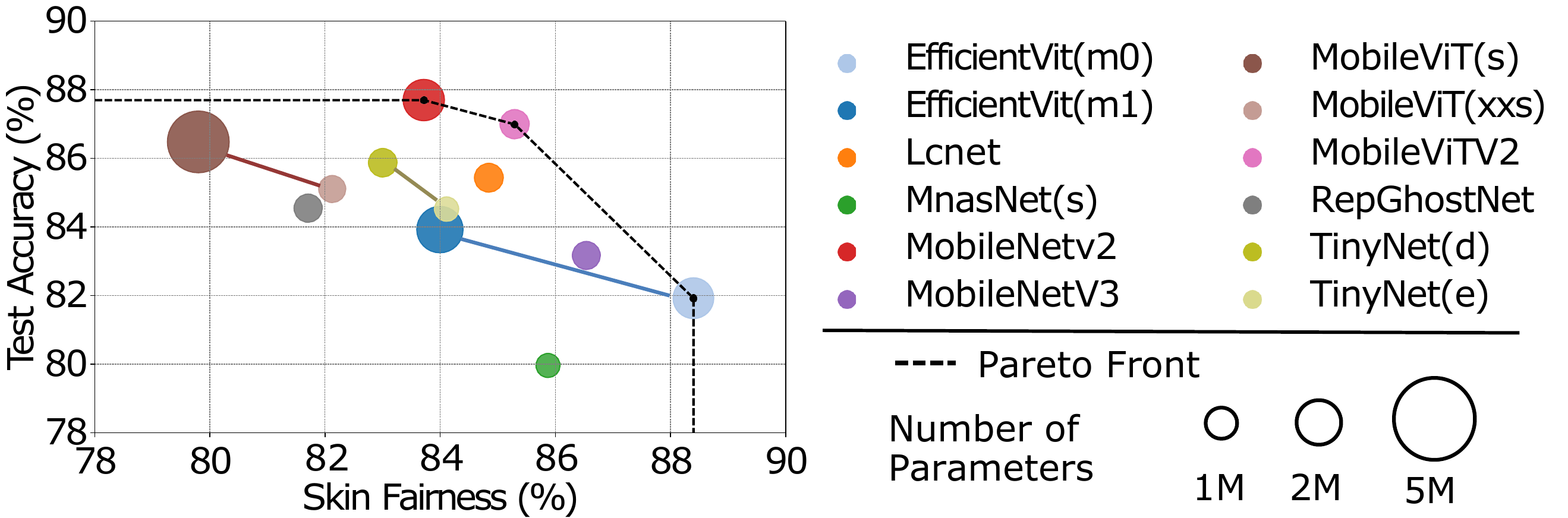}
    \caption{Test Accuracy vs. Skin Fairness of SOTA edge DNNs: Models sharing the same architecture are connected by straight lines. The Pareto front is illustrated with a dashed line. }
    \label{fig:sota_graph}
\end{figure}

Further analysis (Figure~\ref{fig:sota_graph}) demonstrates the influence of model architecture and size on fairness. Models with similar sizes but different architectures exhibit varying fairness scores, indicating that fairness is highly architecture-dependent. Moreover, increasing model size tends to enhance test accuracy but worsens fairness disparities, as larger channel sizes (depth) and more feature representations (width) amplify biases. These findings reveal the need for novel scaling strategies that explicitly target fairness and robustness without solely relying on traditional size scaling techniques.

\subsection{Contributions}

\begin{figure*}[ht]
    \centering \includegraphics[width=.9\linewidth]{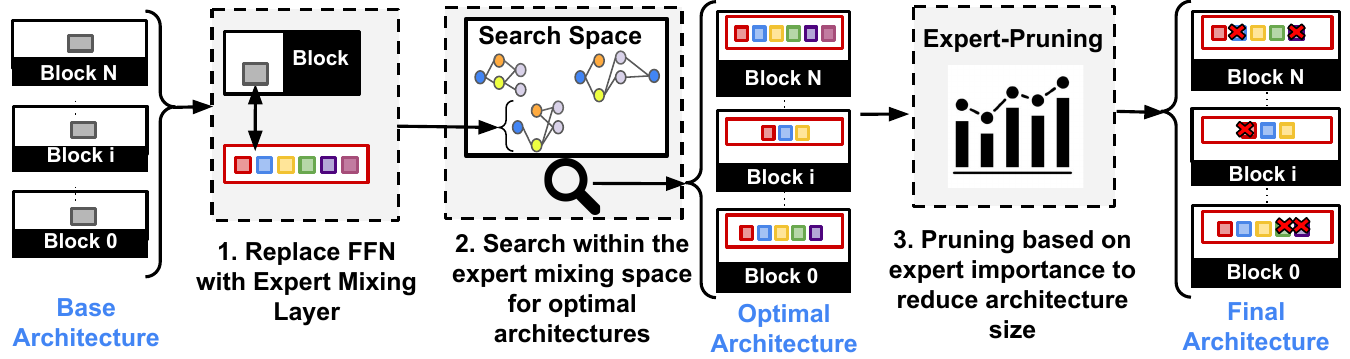}
    \caption{Summary of MoENAS contributions: (1) Replace the FFN layer with a Switch FFN layer, (2) Search within the expert mixing space for optimal architectures (according to accuracy, fairness, robustness, generalization). (3) Pruning based on expert importance for better efficiency. }
    \label{fig:MoENAS_overview}
\end{figure*}

Our key contributions (listed below and summarized in Figure~\ref{fig:MoENAS_overview}), enable the design of fair, robust, and general edge DNNs.
%
%
\begin{enumerate}[leftmargin=*]
    \item \textbf{Dynamic Feature Extraction via Model Scaling}: We propose a scaling approach that varies the number of feature extractors dynamically (inspired by Mixture of Experts (MoE) and Switch layer architectures~\cite{fedus2022switch}), enabling adaptive feature extraction tailored to specific inputs. This flexibility indirectly benefits fairness and robustness by allowing more efficient and context-aware use of resources.
    \item \textbf{HW-NAS with Fairness and Robustness Optimization}: We propose a HW-NAS method with a search space over switching architectures, varying the number of experts per block. Using machine-learning-based performance predictors, the search strategy identifies architectures optimized for accuracy, fairness, robustness, and model size.
    \item \textbf{Expert Pruning for Efficiency}: We introduce an expert pruning method that helps reduce the sizes of models discovered by the search method by iteratively pruning the least used experts. This approach aims to enhance the efficiency of the resulting models without sacrificing performance.
\end{enumerate}
%


\section{Related Works}
\label{sec:related_works}
\subsection{Fairness and Robustness in DNNs}
Significant strides have been made in DNNs, particularly in enhancing fairness, robustness, and generalization. \cite{narayanan2024fairness, ali2024assessing, zhang2021understanding}. Research efforts have predominantly centered around data-centric and algorithmic-centric approaches. Data-centric methods focus on manipulating the input or training data to achieve more balanced representations and outcomes \cite{wu2022fair, celis2020data}, while algorithmic-centric strategies modify the learning algorithms themselves to ensure fairer, more robust model behavior \cite{li2021ditto}. However, very few works delve into the intricate relationship between fairness and robustness and the underlying model design \cite{sheng2022larger}. Moreover, much of the existing literature focuses on large, resource-intensive models which makes their findings inapplicable at the edge \cite{parraga2023fairness, gustafson2023facet}.

\subsection{Hardware Aware Neural Architecture Search}
HW-NAS is a technique that focuses on the automatic design of DNNs. This automation can be presented as a search problem over a set of decisions defining the various components of a DNN aiming to balance and optimize for competing goals such as performance and computational cost\cite{benmeziane2021comprehensive}. HW-NAS has dramatically transformed edge computing, introducing highly efficient models~\cite{terven2023comprehensive,koonce2021mobilenetv3,mehta2021mobilevit}. However, a notable limitation of current HW-NAS efforts, especially in edge computing, is their primary focus on accuracy and efficiency metrics while overlooking other crucial aspects such as fairness, robustness, and generalization, creating a critical gap as demonstrated in Figure~\ref{fig:sota}. \textit{This underscores the need for future NAS methodologies to incorporate these aspects to enhance the design and deployment of equitable and robust edge DNNs.}

\subsection{Mixture of Experts}

The MoE concept is an ensemble learning technique that has been increasingly used in DNNs~\cite{rincy2020ensemble}. It has been successfully applied to scale Natural Language Processing and Computer Vision models ~\cite{du2022glam, riquelme2021scaling}. The Switch Transformer stands out as a prominent example \cite{fedus2022switch}, showcasing an architecture that scales efficiently to a trillion parameters through its use of sparse MoE layers. These layers allows to selectively activate a subset of parameters (or ``experts'') based on the input, significantly enhancing model capacity and performance without a linear increase in computational demand. 


The benefits of MoE extend beyond model scaling, as their use of dynamic routing based on input features indicates potential for improving other metrics such as fairness, robustness, and generalization. The ability of MoE models to adapt based on the inputs makes them versatile and effective in handling complex and varied data, thus enhancing their ability to generalize and maintain good performance across different scenarios (robustness) and various 
subgroups (fairness). \textit{The potential use of MoE (Mixture of Experts) for improving these metrics is still a relatively under-explored area~\cite{cui2022synergy, aimar2023balanced}. However, it represents an intriguing research direction, as demonstrated in this paper.}



\section{\textit{MoENAS} Methodology}
\label{sec:approach}
\subsection{State-of-the-art Analysis: Observations, Challenges, and Goals}

Our methodology is a strategic response to solve the challenges derived from our evaluation of SOTA methods presented in figure~\ref{fig:sota}, The challenges are summarized as follows:
%
\begin{enumerate}[leftmargin=*]
    \item Model designs and architectures directly affect fairness, robustness, and generalization. Therefore, unlike current techniques that concentrate on either enhancing the data quality or tweaking the training algorithms, we aim to enhance these metrics by modifying the model architecture.
    \item Increasing depth (feature size) and width (number of features) of the model has been shown to improve accuracy. but, at the cost of exacerbating fairness disparities. Therefore, We propose an MoE-based scaling approach that can improve performance without compromising fairness.
    \item Existing design techniques for Edge DNNs neglect fairness, robustness, and generalization resulting in biased, non-robust, and non-generalizable models. Therefore, in our work, We embed fairness, robustness, and generalization as objectives or constraints in the Edge DNNs design process.
\end{enumerate}
%
\subsection{Methodology Overview}

\begin{figure}[ht]
    \centering \includegraphics[width=\linewidth]{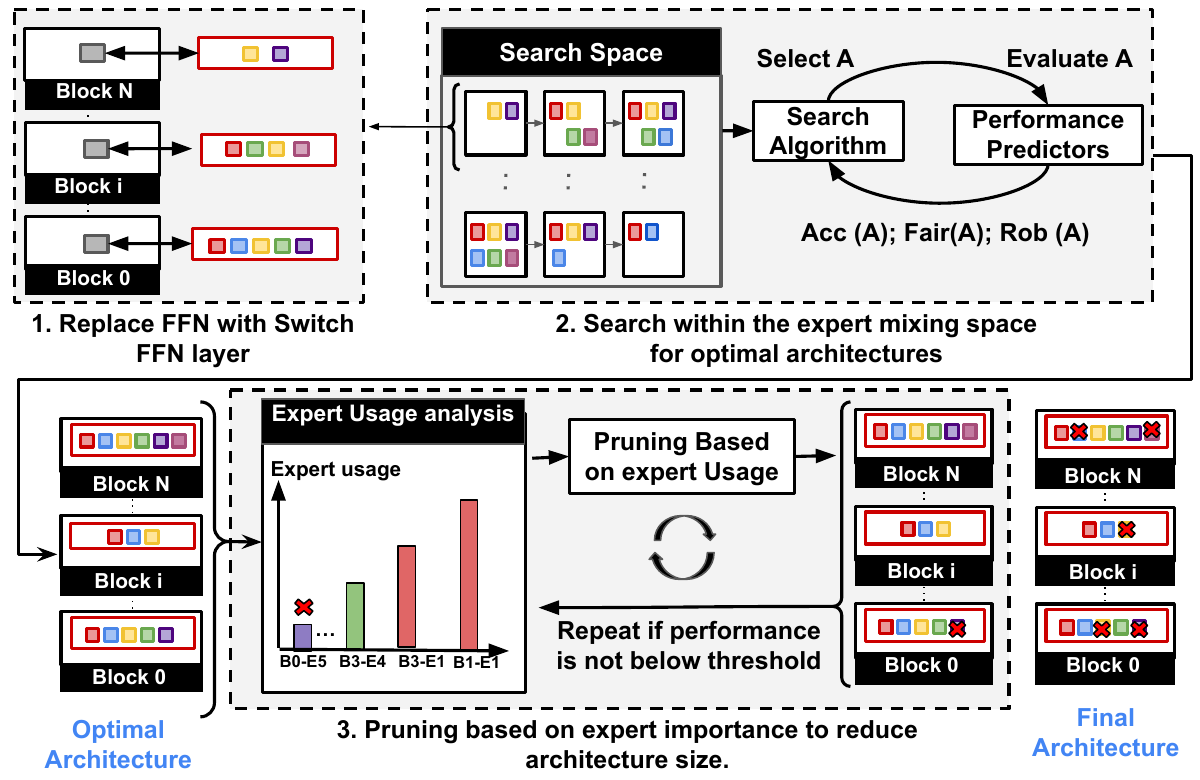}
    \caption{Overview of MoENAS Methodology. This figure illustrates the three core steps of the MoENAS approach: (1) Replace the standard FFN layer (grey) with a Switch FFN layer (colored rectangle refers to the experts) (2) Execution of a search process within the expert mixing space to identify optimal expert combinations for accuracy, fairness, and robustness; (3) Prune the least used experts to ensure model compactness and efficiency while maintaining high performance.}
    \label{fig:MoENAS_details}
\end{figure}

In response to the aforementioned challenges, we introduce MoENAS, a method that utilizes NAS to create edge DNNs that are accurate, fair, robust, and capable of generalization, all while adhering to constraints on model size and efficiency. The MoENAS methodology is structured around three core steps as shown in Figure~\ref{fig:MoENAS_details}: (1) integrating a Switch FNN layer (S-FFN) into an attention-based architecture, (2) search process within the expert mixing space, and (3) expert pruning.

\subsubsection{Switch FFN layer (S-FFN)}
\label{method:switch-fnn}
This step involves replacing the standard FFN layer in the attention block with an S-FFN layer. This modified layer, inspired by \cite{fedus2022switch}, utilizes a MoE technique, enabling dynamic routing of features to different specialized experts. 

\begin{figure}[t]
    \centering
\includegraphics[width=\linewidth]{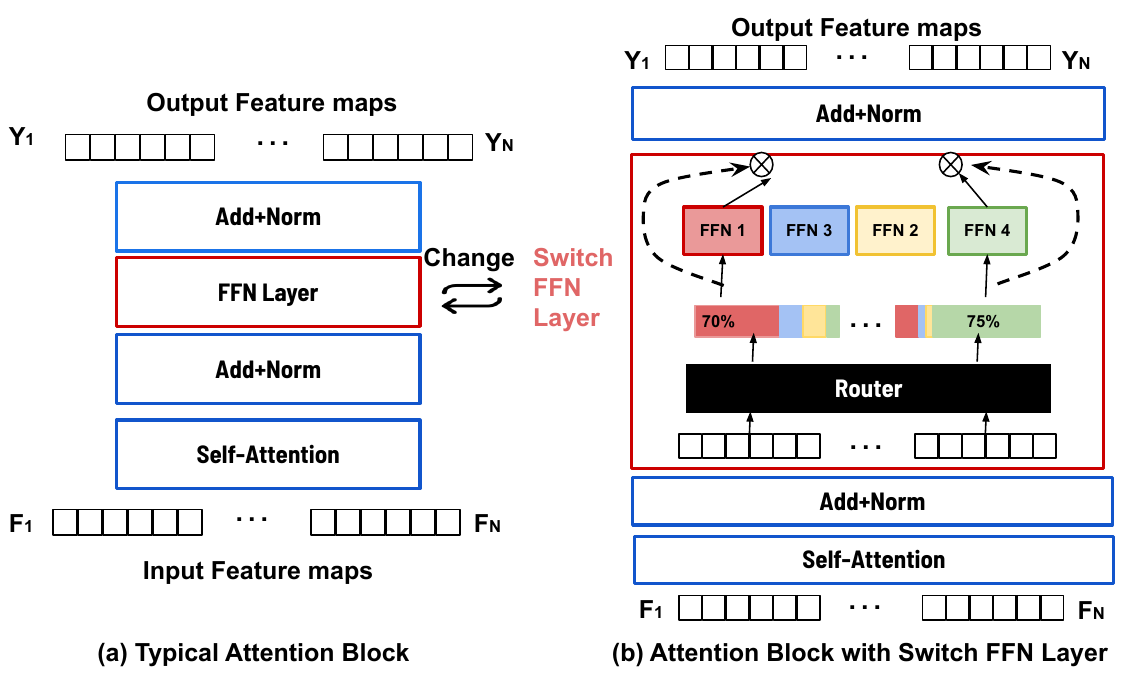}
    \caption{Detailed view of the Attention Block with Switch FFN Layer. This layer is based on MoE, each feature vector is processed by an FFN layer (expert) among the FFN set (experts set) based on the router selection.}
    \label{fig:switch_fnn}
\end{figure}

Traditionally, the structure of the attention block comprises two primary components as shown in Figure~\ref{fig:switch_fnn}.a: an attention layer and a feed-forward network~\cite{vaswani2017attention}. The attention layer's dynamic weight adaptation based on input is a key contributor to its SOTA performance~\cite{khan2022transformers}. However, the static nature of the standard FFN layer, which processes every feature in the same manner, represents a missed opportunity for dynamic adaptation.

Addressing this limitation, we replace the traditional FFN layer with an S-FFN layer. This later selects experts dynamically based on the features using a trained layer called the router. Such an approach ensures that the processing of features is contextually optimized based on their characteristics, thereby extending the dynamic adaptability characteristic of the attention mechanism throughout the model.

\textit{Anatomy of the Switch FFN Layer (S-FFN):} 
The S-FFN layer, as shown in Figure~\ref{fig:switch_fnn}.b, is architecturally composed of two primary components: a router, which directs each feature to a specific expert, and the expert set, where each expert is tailored to process a subset of the feature map based on the router's assignment. This structure allows for an adaptive and efficient processing of features, aligning with our objective to improve fairness, robustness, and generalization in neural network models.
\newline

\textit{Why an MoE-based Layer?}\\
In our NAS strategy, we chose an MoE approach because it allows the model to function like a skilled team of experts, where each expert specializes in analyzing certain features. Just as in a team where each member brings their expertise to focus on the area they know best, MoE models dynamically route inputs to the most appropriate expert. This ensures that each part of the data is processed by the expert most capable of handling it, leading to better overall performance. This adaptability of MoE not only enhances accuracy but also addresses fairness by ensuring that the model's decisions are more equitable across different subgroups.

\subsubsection{Search Process}~\\ \label{method:search-process}
The integration of the S-FFN layer, as previously outlined, presents clear benefits, yet the optimal placement and method of incorporation within the architecture require further exploration. In response, we utilize a NAS strategy to determine the most effective way to include these layers. This approach seeks architectures that achieve a balance between objectives, maximizing accuracy, fairness, and robustness, reducing overfitting, and maintaining a manageable model size.

\begin{figure*}[ht]
    \centering
    \includegraphics[width=.85\linewidth]{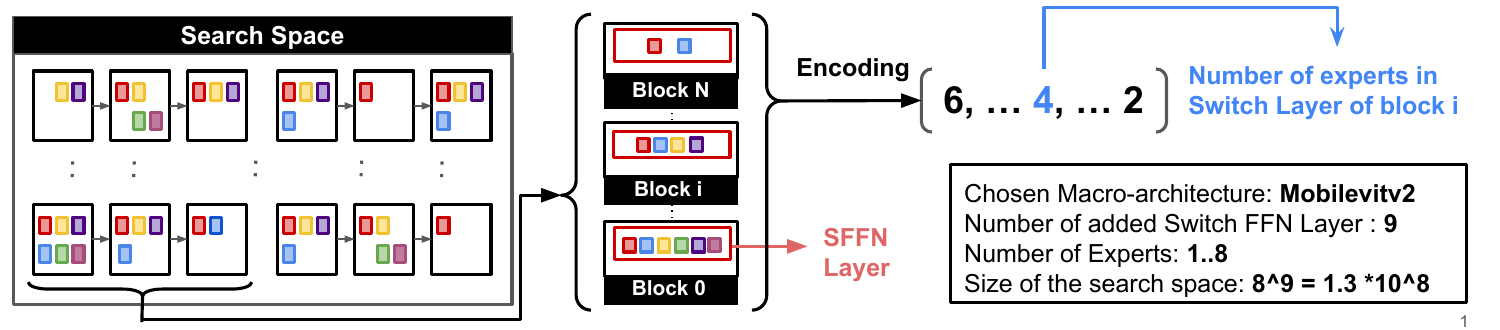}
    \caption{
Search space of MoENAS: Each candidate architecture is built on the MobileViTv2 macro-architecture, varying only in the number of experts per S-FFN layer, represented as a vector encoding the expert count for each S-FFN layer.}
    \label{fig:search_space}
\end{figure*}

The proposed search process begins with the selection of an attention-based architecture, for which we chose MobileNetVitV2, based on experimental analyses(see Figure~\ref{fig:sota}). Within this model, all FFN layers are substituted with Switch FFN layers as demonstrated in Figure~\ref{fig:switch_fnn}. The variation in the number of experts within each Switch FFN layer constitutes the core of our search space diversity. Specifically, in the case of MobileNetVitV2, which comprises nine layers, each layer can have from 1 to 8 experts. Consequently, each architecture configuration within this search space is encoded as a vector, succinctly representing the number of experts in each layer as shown in Figure~\ref{fig:search_space}.

To explore this space, we employed a modified Bayesian Optimization (BO) strategy \cite{white2021bananas}, optimized to concurrently address four objectives: Test Accuracy, Skin Fairness, Robustness, and Generalization, while treating model size as a constraint. This approach utilized a population-based BO mechanism\cite{pelikan2002scalability}. To guide the search, we train surrogate models capable of estimating each objective metric from the vector representation of an architecture as shown in Figure~\ref{fig:search_space}. Here, XGBoost was selected as the surrogate, known for its effectiveness in handling such input vectors \cite{benmeziane2021comprehensive}. 

\subsubsection{Expert Pruning}~\\ \label{method:expert-pruning}
%
While incorporating the Switch FFN Layer in place of the standard FFN does not impact the computational cost (measured in FLOPs), the model's size increases due to the presence of multiple FFNs within the switch layer. To manage this increase, we have carefully designed our search space to ensure all models remain within a manageable size range of 1M to 3M parameters. To further optimize the model size, we introduce a strategy called expert pruning inspired by \cite{chen2022task}.

Expert pruning (Figure~\ref{fig:MoENAS_overview}), is a method aimed at reducing the model's size by removing the least utilized experts. First, we run the model across all validation images to track the frequency of each expert's use. Then, we rank the experts based on their usage and identify those that are least utilized and therefore are candidates for pruning. 

Once an expert is pruned, the features that would have been routed to it are redirected to their next best option, as determined by the router's probabilities. Following this, the model undergoes evaluation to assess its performance. If the performance remains above a predetermined threshold, the process can be repeated to further refine the model's size and efficiency. This iterative approach ensures that we maintain a balance between model size, efficiency, and performance, making our models efficient for deployment in resource-constrained edge computing environments.

\textit{A question that may arise is why we use post-search pruning when pruned models are already in the search space.} The key reason is that post-search pruning occurs after the model has been fully trained, similar to how knowledge distillation works. Just as a smaller model distilled from a larger one outperforms the same model trained from scratch, training a larger MoE model and then pruning it can yield better results than directly training a smaller model. This post-training adjustment allows us to optimize the model, preserving and enhancing the most effective pathways identified during training, ultimately leading to improved efficiency.

\label{sec:metrics}
\subsection{Evaluation Metrics}
To assess the performance of our models, we focus on several key metrics: 
\begin{itemize}[leftmargin=*]
    \item \textbf{Validation and Test Accuracy:} This metric evaluates the model's accuracy on person classification using COCO~\cite{lin2014microsoft} for the validation and FACET~\cite{lin2014microsoft} for the test.
    \item \textbf{Skin Fairness:} examine the consistency of model accuracy across different skin tones. To calculate it, we adopt a formula that adjusts the Statistical Parity Difference (SPD)\cite{sheng2022larger}, used to calculate unfairness, with a constant $\beta$, to quantify fairness. The formula is given by:
     \begin{equation}
    \text{Fairness} = \frac{\beta - \text{SPD}}{\beta}
    \end{equation}
    \begin{equation}
    \text{SPD} = \sum_{i=1}^{N} \left| \text{Acc}_{G_i} -
    \text{Acc}_{mino} 
    \right|
    \label{eq:spd}
\end{equation}
%
%
    $\beta$ is set to 0.2. $N$ is the number of groups (10 skin tones), and $\text{Acc}_{G_i}$ is the accuracy for group $i$, with $\text{Acc}_{\text{mino}}$ being the accuracy of the minority group (the group with the least number of images).
    \item \textbf{Robustness to Light:} Measures the model's performance consistency under poor lighting conditions by calculating the accuracy for the subset of poorly lit images in the test dataset.
\item \textbf{Overfitting:} Measured by the difference between validation and test accuracy, it indicates the model's generalization ability. A smaller gap suggests better generalization to unseen data since test and training/validation data come from two different distributions.
\end{itemize}

\section{Evaluation of our \textit{MoENAS} Methodology}
\label{sec:eval}
In the experimental section, we evaluate our method by benchmarking it against state-of-the-art edge DNNs, selected from the Hugging Face list of the fastest and most accurate edge models~\cite{huggingface_timm_fastest}. Before the comparative analysis, we outline the key elements of our experiments: the datasets used and the training schemes of the models under consideration.

\subsection{Datasets}
\label{sec:dataset}
To train and test the models, we strategically employ two datasets, COCO and FACET, each chosen for their suitability to different aspects of our study. The COCO~\cite{lin2014microsoft} dataset is utilized for both training and validation, And the FACET dataset~\cite{gustafson2023facet} is used for testing. 

Facet includes exhaustively labeled, manually annotated attributes covering demographic and physical traits of individuals across 32k images, allowing for a fine-grained analysis of model performance across diverse demographic groups. Unlike many benchmarks that simplify attributes such as skin tone into binary categories like white or black, FACET employs a nuanced approach by using a spectrum from 1 to 10 to represent varying skin tones, capturing subtle differences in skin color. This detailed annotation system, combined with the comprehensive scope of the dataset, makes FACET the best choice for assessing and improving the fairness and robustness of vision models across multiple demographic axes~\cite{gustafson2023facet}.

\begin{table*}[ht]
\footnotesize
\centering
\caption{Comparison of model performance: MoENAS vs SOTA edge DNNS}
\label{tab:results}
\begin{adjustbox}{max width=\linewidth}
\begin{tabular}{|p{2.7cm}|p{5.6cm}|>{\centering\arraybackslash}p{1.2cm}|>{\centering\arraybackslash}p{1.2cm}|>{\centering\arraybackslash}p{1.3cm}|>{\centering\arraybackslash}p{1.3cm}|>{\centering\arraybackslash}p{1.1cm}|>{\centering\arraybackslash}p{1.1cm}|>{\centering\arraybackslash}p{1.3cm}|}
\hline
\textbf{Category} & {\makecell{\\ \\ \textbf{Model}}} 
& \multicolumn{1}{c|}{\multirow{3}{*}{\makecell{\textbf{Test} \\ \textbf{Accuracy} \\ (\%)}}}  
& \multicolumn{3}{c|}{\makecell{ \textbf{Fairness}}} 
& \multicolumn{1}{c|}{\multirow{2}{*}{\makecell{\textbf{Robust-} \\ \textbf{ness} (\%)}}}   
& \multicolumn{1}{c|}{\multirow{2}{*}{\makecell{\textbf{Overfit-} \\ \textbf{ting}(\%)}}} 
&  \multicolumn{1}{c|}{\multirow{3}{*}{ \makecell{\textbf{Model size} \\ (Million \\ Params)}}}
\\ \cline{4-6}
& & & \makecell{\textbf{Fairness} \\ \textbf{Score} \\ (\%)}  & \makecell{\textbf{Accuracy} \\ \textbf{Lightest} \\ \textbf{Skin}(\%)} & \makecell{\textbf{Accuracy} \\ \textbf{Darkest} \\ \textbf{Skin}(\%)} &  & & \\ 
\hline 

 & EfficientVit(m0)~\cite{cai2023efficientvit} & 81.92 & 83.08 & 89.14 & 78.68 & 76.31 & 8.78 & 2.16 \\

& EfficientVit(m1)~\cite{cai2023efficientvit} & 85.69 & 85.79 & 92.84 & 84.16 & 80.29 & 4.05 & 2.79 \\

 & RepGhostNet~\cite{chen2022repghost} & 84.55 & 79.49 & 91.88 & 80.21 & 79.76 & 7.48 & 1.04 \\

 & MobileVit(xxs)~\cite{mehta2021mobilevit} & 85.11 & 82.13 & 91.82 & 82.70 & 76.31 & 7.34 & 0.95 \\

\textbf{Edge SOTA models}& LcNet~\cite{cui2021pp} & 85.44 & 84.85 & 92.4 & 83.58 & 82.14 & 6.60 & 1.08 \\

& MobileVit(s)~\cite{mehta2021mobilevit} & 86.48 & 79.80 & 92.78 & 83.58 & 78.29 & 5.78 & 4.94 \\

& MobileVitV2~\cite{mehta2022separable} & 86.99 & 85.29 & 93.23 & 85.04 & 82.72 & 6.50 & 1.11 \\

& MobileNetV2~\cite{sandler2018mobilenetv2} & 87.70 & 79.51 & 95.19 & 84.17 & 83.54 & 4.62 & 2.23 \\

& Resnet18~\cite{he2016deep} & 88.05 & 77.02 & 94.12 & 83.87 & 83.93 & 7.21 & 11.17 \\

\hline

& MnasNet(s)~\cite{tan2019mnasnet} & \centering 81.81 & \centering 77.00 & 88.37 &  78.00  
& 71.33 & 6.88 & \colorbox{lime}{\textbf{0.75}} \\

& MobileNetV3(s)~\cite{howard2019searching} & 82.85 & 83.33 & 91.25 & 80.65 & 77.34 & 7.71 & 1.02 \\

\textbf{HW-NAS models} & Tinynet(e)~\cite{han2020model} & 85.38 & 80.44 & 92.84 & 82.4 & 79.81 & 6.99 & 0.76 \\

& TinyNet(d)~\cite{han2020model} & 86.05 & 86.63 & 92.52 & 85.04 & 81.56 & 6.63 & 1.06 \\

 & ProxylessNas~\cite{cai2018proxylessnas} & 87.39 & 76.21 & 92.65 & 83.28 & 82.44 & 5.82 & 5.39 \\

\hline

 & \textbf{MoENAS-S} & 90.33 & \colorbox{lime}{\textbf{93.13}} & 95.59 & \colorbox{lime}{\textbf{90.62}} & 84.51 & \colorbox{lime}{\textbf{0.21}} & 2.71 \\
\textbf{Proposed Models} & \textbf{MoENAS-XS} & 89.4 & 87.51 & 94.38 & 87.68 &  85.54 & 0.74 & 2.43 \\
& \textbf{MoENAS-XXS} & \colorbox{lime}{\textbf{91.72}} & 88.27 & \colorbox{lime}{\textbf{96.55}} & 88.27 & \colorbox{lime}{\textbf{90.32}} & 3.79 & 1.90 \\
\hline
\end{tabular}
\end{adjustbox}
\end{table*}

\subsection{Training Setup}

To ensure a fair and comprehensive comparison, each SOTA model is initialized with its pre-trained backbone on the ImageNet dataset~\cite{deng2009imagenet}. To accommodate the diverse optimization strategies inherent to each model, we employ five distinct training schemes. These schemes are derived from the best practices recommended in the respective papers of the SOTA models~\cite{sandler2018mobilenetv2, mehta2021mobilevit, tan2019efficientnet}. For each model, we conduct training sessions under each of the five schemes and evaluate their performance based on test accuracy. The scheme that yields the highest test accuracy for a model is then selected as the optimal training approach for that particular model.

The models from our search space are trained only on one chosen scheme among the five, which is the most effective for MobileVitV2~\cite{mehta2022separable} since it is our base architecture. The training duration for each model, including our own, spans 150 epochs, allowing sufficient time for the models to converge.

\subsection{Hardware and Computational Cost}
\label{app:hw}
Training was conducted on an internal machine server equipped with 64 CPU cores, 4 A6000 GPUs, and 256 GB of RAM. The search cost is primarily dominated by the time required to train models from the search space in order to train the surrogate models. A total of 127 models were trained, each taking an average time of 2 hours and 21 minutes per model. The total GPU hours required for training all models amount to approximately 299 hours (2.35 hours per model * 127 models).

However, due to the availability of 4 GPUs and the capability to train 4 models simultaneously, the actual search time was significantly reduced. The real search time was about 5 days (114 hours), allowing us to efficiently complete the training process.

\subsection{Results and Discussion}
\subsubsection{Performance Across Metrics}
The results of our experimental analysis are reported in Table~\ref{tab:results}. Notably, our MoENAS models exhibit superior performance across all key metrics (Test Accuracy, Skin Fairness, Robustness, Overfitting) compared to the SOTA edge DNNs. Specifically, \mbox{MoENAS-S} achieves the highest performance on Skin Fairness (93.13\%). MoENAS-XXS not only excels in Test Accuracy (91.72\%), but also showcases good robustness (87.34\%) and low overfitting (3.79\%). 
These results illustrate the model's ability to generalize across diverse conditions without compromising overall performance.

\subsubsection{Performance Across Skin Tones}

%
%
To further demonstrate the performance of our MoENAS-S model, we compare its accuracy across 10 different skin tones against the average accuracy of SOTA as reported in Figure~\ref{fig:Accuracy_Skin_Tone_Models} (a). Our findings reveal that the MoENAS-S model not only enhances the accuracy for each skin tone but also significantly narrows the accuracy gap between the lightest and darkest skin tones ($\Delta$ of 5.6\%), in contrast to the observed average of the SOTA models ($\Delta$ of 14.1\%).


These results demonstrate our model's capability to uplift overall performance while concurrently minimizing disparities across skin tones, thus successfully addressing the gaps identified in the motivation section of our study. 

\subsubsection{Detailed Comparative Analysis with State-of-the-Art Models}
\label{app:one-one}
\begin{figure*}[ht]
    \centering
    \includegraphics[width=\linewidth]{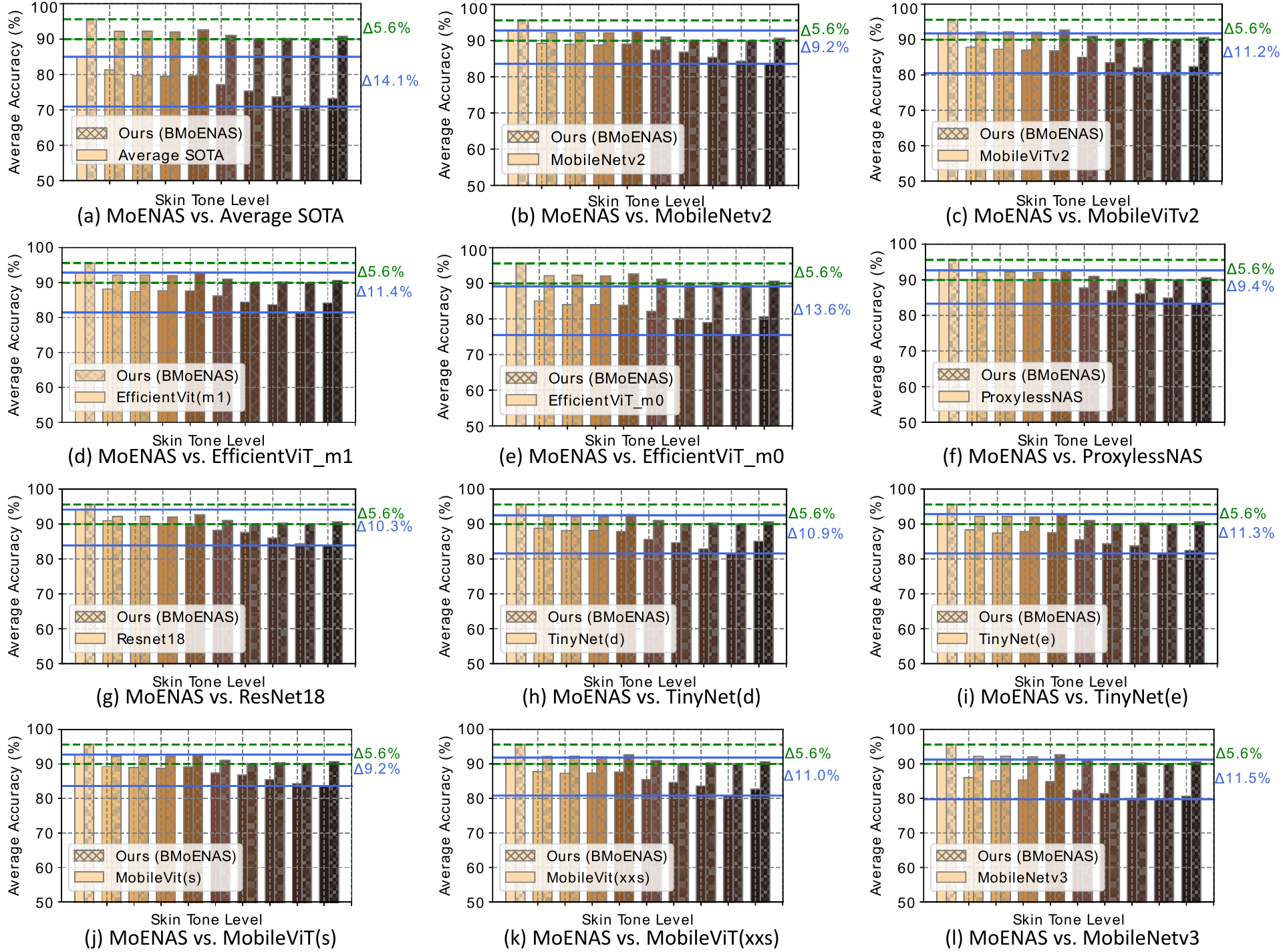}
    \caption{Accuracy across Skin Tones. (a) MoENAS vs. Average SOTA models. (b-l) MoENAS vs. Individual SOTA models. MoENAS achieves better accuracy for all individual skin tones while also reducing the gap the accuracy gap between the lightest and darkest skin tones to $\Delta$ = 5.6\%.}
    \label{fig:Accuracy_Skin_Tone_Models}
\end{figure*}

We also provide a detailed one-on-one comparison of the MoENAS-S model against 10 state-of-the-art (SOTA) models as shown in Figure~\ref{fig:Accuracy_Skin_Tone_Models} (b-l). The MoENAS-S model consistently outperformed each of the 10 SOTA models in terms of accuracy for all 10 skin tones. Also, MoENAS-S model achieved a gap of only 5.6\%, while the lowest gap achieved by the SOTA models is 9.2\%. This substantial reduction highlights our model's effectiveness in addressing disparities across different skin tones and therefore reducing bias. 

\subsubsection{Performance and Pareto Front Analysis}
\label{app:pareto}
\begin{figure*}[ht]
    \centering
    \includegraphics[width=.9\linewidth]{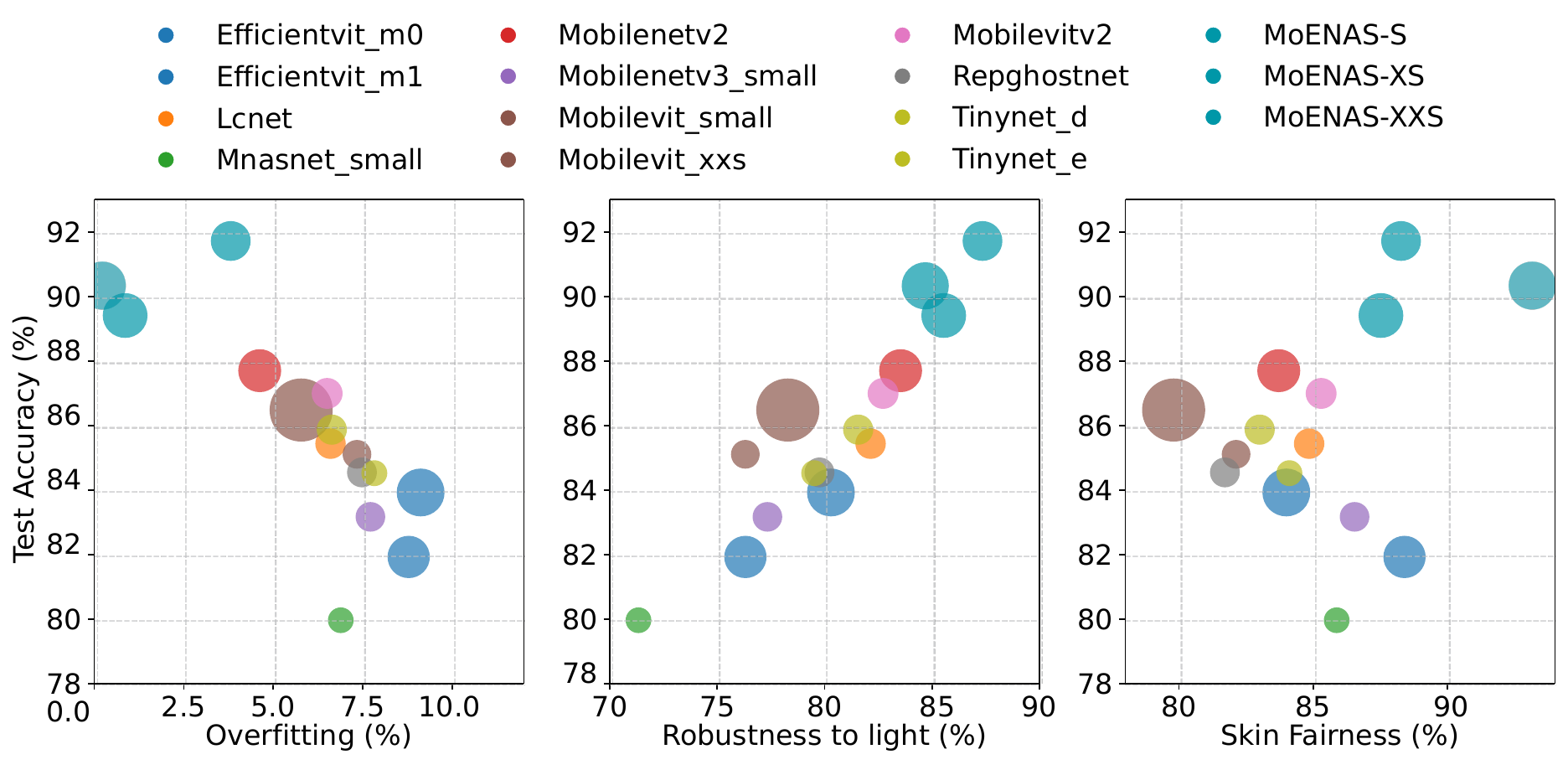}
    \caption{Performance comparison of MoENAS vs SOTA based on Test Accuracy, Overfitting, Robustness, skin fairness, and model size. The graphs show that our models (in green) dominate SOTA models on all metrics.}
    \label{fig:results}
\end{figure*}

The results from Table \ref{tab:results} demonstrate that our MoENAS models consistently outperform the previous SOTA Pareto front across various metrics. This superiority is visually evident in Figure \ref{fig:results}, where our models are positioned on the Pareto front for all three graphs. These findings indicate that MoENAS models achieve an efficient balance between accuracy, fairness, robustness, and compactness, making them ideal for edge computing applications. In summary, MoENAS models not only set new benchmarks but also excel in designing edge DNNs that are powerful, efficient, and fair for real-world applications.

\subsubsection{Discussion}
In summary, our proposed MoENAS method demonstrates an effective approach to bridge the gaps in the performance of edge DNNs with respect to fairness, robustness, and generalization. 
 Our methodology effectively creates more equitable, robust, and efficient edge DNNs, advancing the state of edge DNN design. However, MoENAS slightly increases the model size by 0.4M compared to the average edge DNNs' sizes. This trade-off opens new avenues for further optimization. Future work could focus on minimizing this increase in size while also expanding the model's objectives to include factors like latency or explainability, thereby broadening the scope and application of our approach in edge computing environments.

\section{Ablation Studies}
\label{sec:study}

\subsection{Expert Pruning}

In the ablation study, we analyze how pruning experts affects the performance and size of the model providing insights into the trade-offs between model complexity and accuracy. 

\begin{figure}[!ht]
\centering
    \includegraphics[width=.9\linewidth]{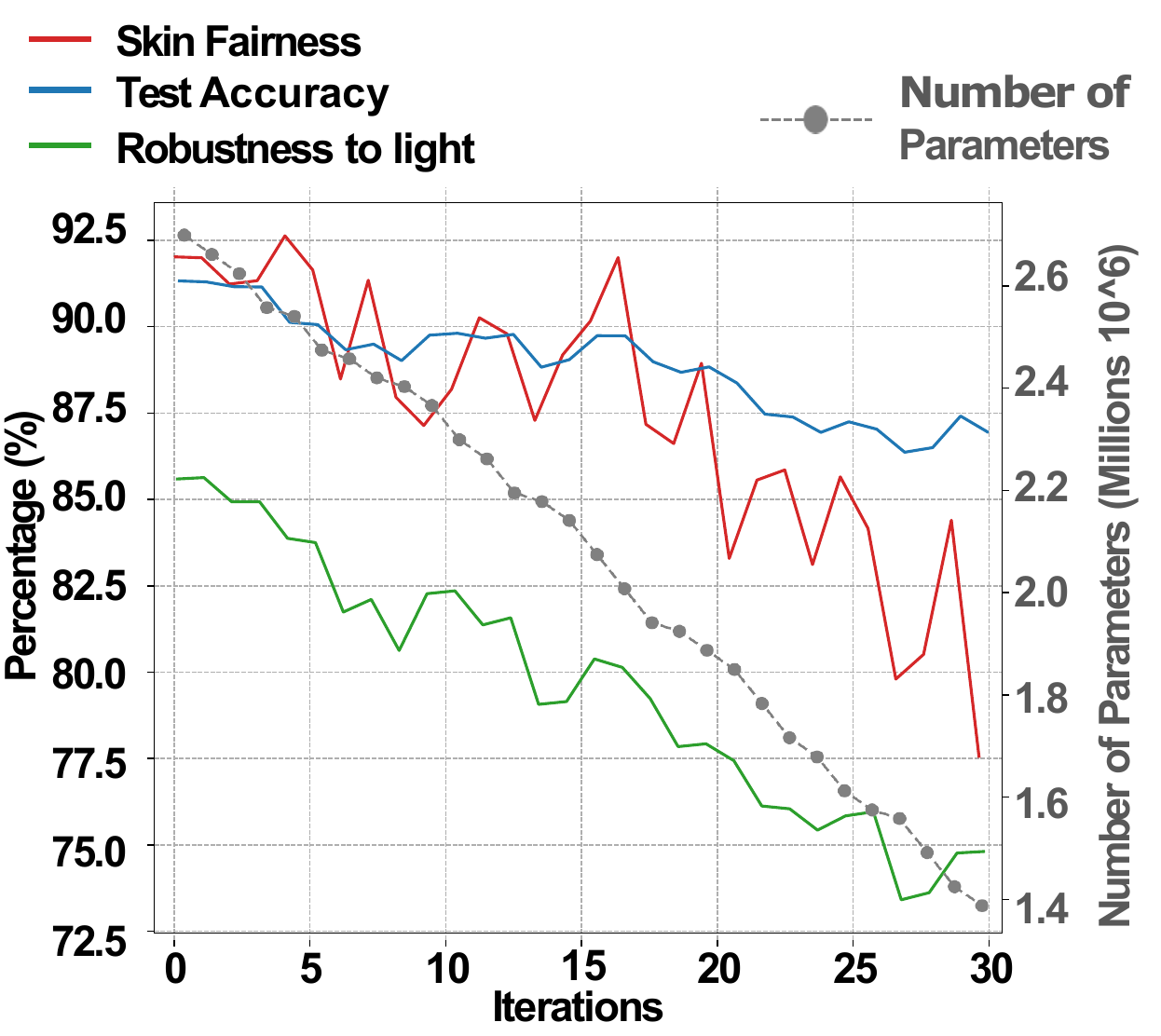}
        \caption{Evolution of test accuracy, robustness, skin fairness, and model size w.r.t. the number of iterations.}
        \label{fig:res_plot_ep_b}
\end{figure}

This study reveals the results of our expert pruning approach. Figure~\ref{fig:res_plot_ep_b} reports the evolution of key metrics (test accuracy, robustness, skin fairness, and model size) w.r.t. the number of iterations, while in each iteration the least used expert is pruned. Up to 16 iterations, The model maintains better accuracy and fairness than SOTA models with a 26\% decrease in model size (from 2.7M to 2.0M) at the cost of a slight 2\% decrease in Robustness. These findings underscore the effectiveness of our pruning method for efficiently reducing model size while keeping key performance metrics above predefined thresholds.

\subsection{Effect of Expert Choice on the Results}
\label{app:expert}
\begin{figure}[ht]
    \centering
    \includegraphics[width=\linewidth]{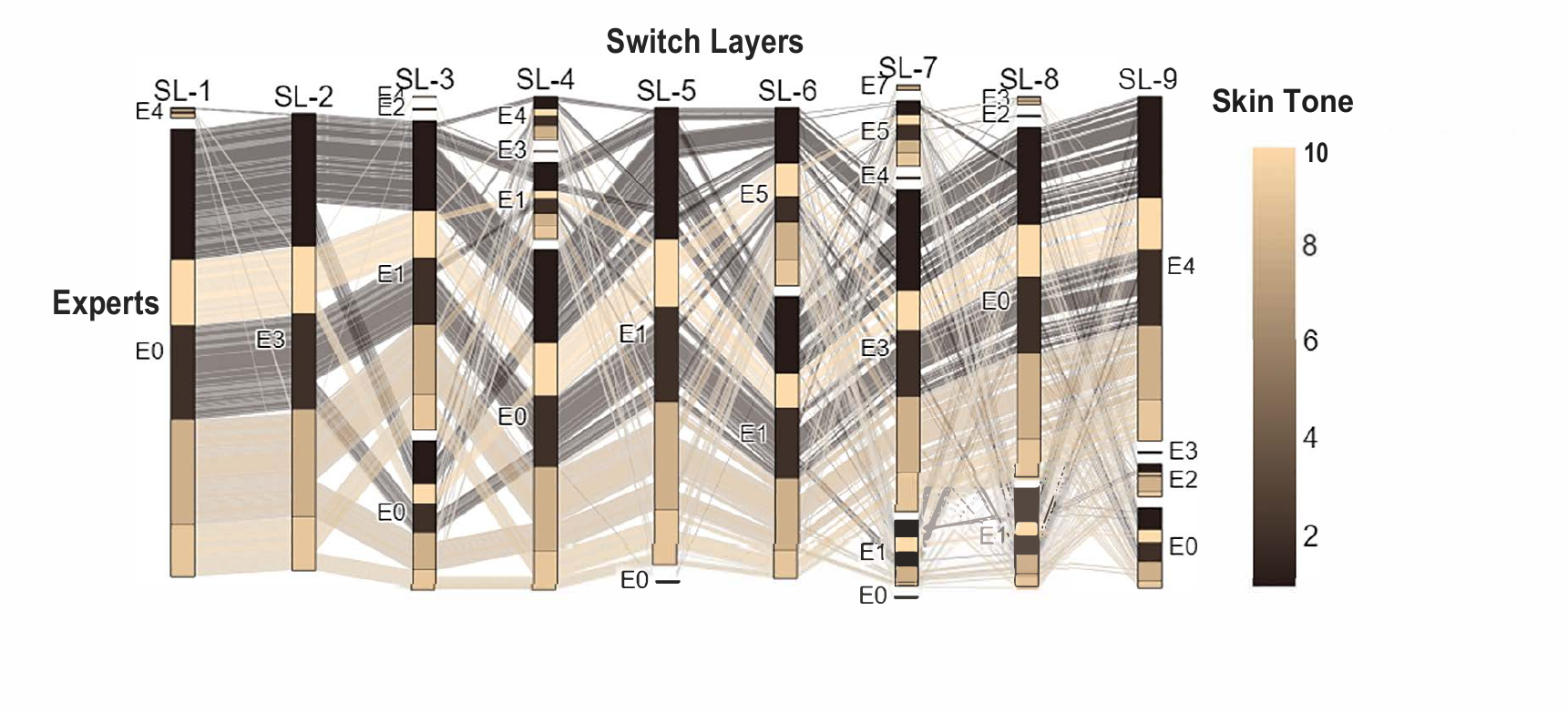}
    \caption{This figure illustrates the expert selection patterns within different switch layers of our model. Each line represents the choice of a specific expert for an individual image across various layers. ``SL-1" refers to Switch Layer 1, and ``E1" denotes Expert 1 in that switch layer, highlighting how images from similar skin tone groups tend to follow comparable paths through the network.}
    \label{fig:expert_choice}
\end{figure}

In our ablation study, we analyzed the performance of our model by examining the choice of experts for each image across different skin tone groups. The results are presented in Figure~\ref{fig:expert_choice}, where each line represents the choice of an expert at each layer for one image.

We observed that clusters of the same color are forming in each layer, suggesting that images from the same skin tone group tend to follow the same path (i.e., they have similar expert choices). Additionally, we noticed that in each layer, there are two or at most three main selected experts, while others are chosen less frequently. This likely stems from the fact that most cases can be effectively handled by these main experts. However, for some extreme cases where the primary experts may fail, the image is routed to a more specific expert.

In summary, thanks to the Switching Layer, the model performs a form of clustering for images, allowing it to choose the appropriate expert for each cluster of images.

\subsection{Impact of a Quantum Head on Edge DNN Architecture Performance}
\label{app:quantum}
\begin{figure}
        \centering
\includegraphics[width=\linewidth]{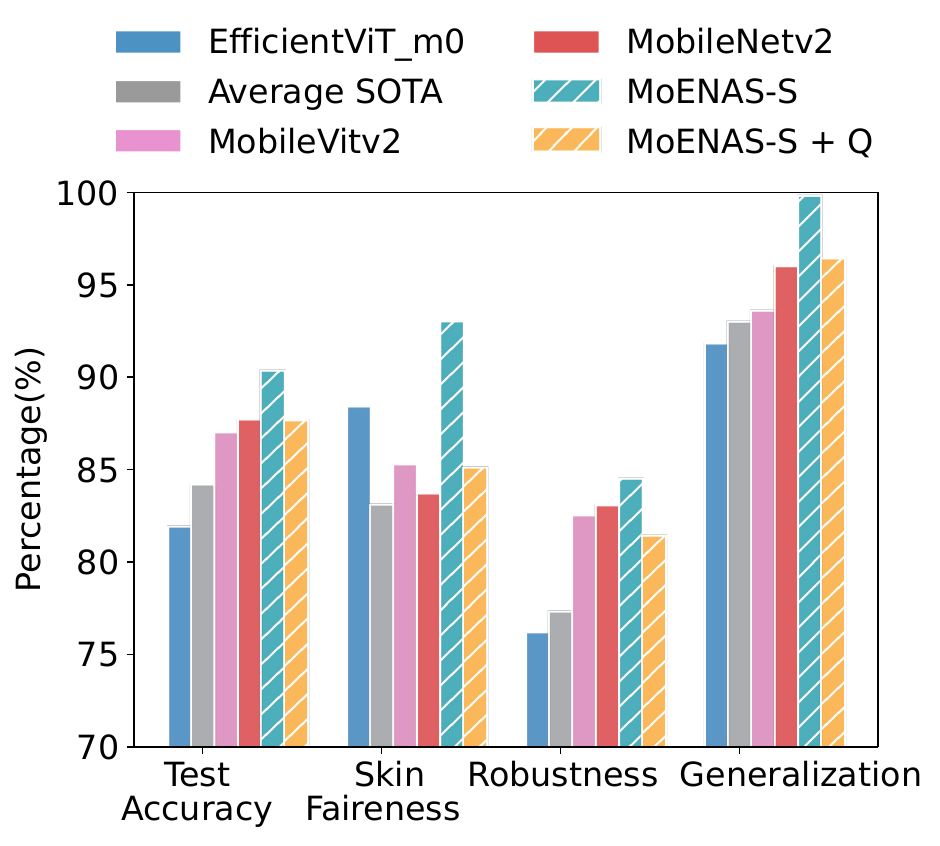}
        \caption{Variation of test accuracy, skin fairness, robustness, and generalization, using SOTA models, MoENAS-S, and our proposed model with quantum heads (MoENAS-S + Q).}
        
\label{fig:res_plot_qh_}
\end{figure}

In this analysis, we delved into the potential of merging classical and quantum computing within edge computing frameworks, motivated by the evolving landscape towards a symbiotic classical-quantum integration\cite{furutanpey2023architectural}. We replaced the classical head of our discovered MoENAS-S architecture with a quantum counterpart. As shown in Figure~\ref{fig:res_plot_qh_}, the resulting model, called \mbox{MoENAS-S + Q}, shows a decline in key performance indicators such as accuracy, fairness, robustness, and generalization when compared to MoENAS-S. However, \mbox{MoENAS-S + Q} surpassed existing SOTA models overall, albeit with a 2\% decrease in robustness. This highlights the viability and promise of hybrid classical-quantum computing approaches in edge DNNs.

\section{Limitations}
\label{sec:Limitations}

Despite the clear advantages in terms of accuracy, fairness, and robustness of our MoENAS framework, it has some limitations. the main one is regarding the model size, due to the nature of MoEs, the model size cannot be reduced as much as standalone edge DNNs, which limits the applicability of MoENAS to moderately low resources, while it is unsuitable for extremely tiny ML use cases. On the other hand, this paper has obvious positive societal impacts that improve fairness in DNNs.

\section{Conclusion}
Our paper introduces Mixture-of-Experts-based Neural Architecture Search (\textit{MoENAS}), a novel approach that goes beyond enhancing accuracy and efficiency of edge DNNs by also addressing other critical issues such as fairness, robustness, and generalization. By utilizing a mixture of experts, \textit{MoENAS} has significantly improved accuracy by 4.02\%, skin fairness by 4.61\%, robustness by 3.80\%, and minimized overfitting to 0.21\%, all while maintaining model size close to the average SOTA size. Future research directions include applying our method to other attention-based models and extending its capabilities to convolutional networks, aiming for broader applicability. Another promising research direction is investigating the explainability of \textit{MoENAS} by analyzing the choice of experts within the architecture. This approach could provide deeper insights into the decision-making process of the method and the network, enabling the development of more transparent and interpretable NAS methods and edge DNNs.

\section*{Acknowledgment}
This work was supported in part by the NYUAD Center for Quantum and Topological Systems (CQTS), funded by Tamkeen under the NYUAD Research Institute grant CG008.

\bibliographystyle{ieeetr}
\bibliography{main.bib}

\begin{thebibliography}{10}

\bibitem{yang2021intelligent}
S.~Yang, F.~Zhu, X.~Ling, Q.~Liu, and P.~Zhao, ``Intelligent health care: Applications of deep learning in computational medicine,'' {\em Frontiers in Genetics}, 2021.

\bibitem{macas2022survey}
M.~Macas, C.~Wu, and W.~Fuertes, ``A survey on deep learning for cybersecurity: Progress, challenges, and opportunities,'' {\em Computer Networks}, 2022.

\bibitem{grigorescu2020survey}
S.~Grigorescu, B.~Trasnea, T.~Cocias, and G.~Macesanu, ``A survey of deep learning techniques for autonomous driving,'' {\em Journal of Field Robotics}, 2020.

\bibitem{alvear2023edge}
V.~Alvear-Puertas, P.~D. Rosero-Montalvo, V.~F{\'e}lix-L{\'o}pez, and D.~H. Peluffo-Ord{\'o}{\~n}ez, ``Edge artificial intelligence for internet of things devices: Open challenges,'' {\em DITTET}, 2023.

\bibitem{Marchisio_2018IJCNN_PruNet}
A.~Marchisio, M.~A. Hanif, M.~Martina, and M.~Shafique, ``Prunet: Class-blind pruning method for deep neural networks,'' in {\em IJCNN}, 2018.

\bibitem{vadera2021methods}
S.~Vadera and S.~Ameen, ``Methods for pruning deep neural networks,'' 2021.

\bibitem{Hanif_2022MICPRO_EfficientEmbeddedDL}
M.~A. Hanif and M.~Shafique, ``A cross-layer approach towards developing efficient embedded deep learning systems,'' {\em Microprocess. Microsystems}, 2022.

\bibitem{chen2021quantization}
W.~Chen {\em et~al.}, ``Quantization of deep neural networks for accurate edge computing,'' {\em JETC}, 2021.

\bibitem{matsubara2020head}
Y.~Matsubara {\em et~al.}, ``Head network distillation: Splitting distilled deep neural networks for resource-constrained edge computing systems,'' {\em IEEE Access}, 2020.

\bibitem{benmeziane2021comprehensive}
H.~Benmeziane {\em et~al.}, ``A comprehensive survey on hardware-aware neural architecture search,'' {\em arXiv preprint arXiv:2101.09336}, 2021.

\bibitem{zhang2020fast}
L.~L. Zhang {\em et~al.}, ``Fast hardware-aware neural architecture search,'' in {\em CVPR Workshops}, 2020.

\bibitem{he2021automl}
X.~He, K.~Zhao, and X.~Chu, ``Automl: A survey of the state-of-the-art,'' {\em Knowledge-Based Systems}, 2021.

\bibitem{elsken2019neural}
T.~Elsken, J.~H. Metzen, and F.~Hutter, ``Neural architecture search: A survey,'' {\em JMLR}, 2019.

\bibitem{wistuba2019survey}
M.~Wistuba, A.~Rawat, and T.~Pedapati, ``A survey on neural architecture search,'' {\em arXiv preprint arXiv:1905.01392}, 2019.

\bibitem{Marchisio_2020ICCAD_NASCaps}
A.~Marchisio {\em et~al.}, ``Nascaps: {A} framework for neural architecture search to optimize the accuracy and hardware efficiency of convolutional capsule networks,'' in {\em {ICCAD}}, 2020.

\bibitem{Prabakaran_2021JIOT_BioNetExplorer}
B.~S. Prabakaran {\em et~al.}, ``Bionetexplorer: Architecture-space exploration of biosignal processing deep neural networks for wearables,'' {\em {IEEE} IoT}, 2021.

\bibitem{Kaisa1999MultiObjective}
M.~Kaisa, {\em Nonlinear Multiobjective Optimization}.
\newblock International Series in Operations Research \& Management Science, Kluwer Academic Publishers, 1999.

\bibitem{sheng2022larger}
Y.~Sheng {\em et~al.}, ``The larger the fairer? small neural networks can achieve fairness for edge devices,'' in {\em DAC}, 2022.

\bibitem{drenkow2021systematic}
N.~Drenkow, N.~Sani, I.~Shpitser, and M.~Unberath, ``A systematic review of robustness in deep learning for computer vision: Mind the gap?,'' {\em arXiv preprint arXiv:2112.00639}, 2021.

\bibitem{porrello2020robust}
A.~Porrello, L.~Bergamini, and S.~Calderara, ``Robust re-identification by multiple views knowledge distillation,'' in {\em ECCV}, 2020.

\bibitem{zhou2022domain}
K.~Zhou {\em et~al.}, ``Domain generalization: A survey,'' {\em TPAMI}, 2022.

\bibitem{esteva2021deep}
A.~Esteva {\em et~al.}, ``Deep learning-enabled medical computer vision,'' {\em NPJ digital medicine}, 2021.

\bibitem{mehrabi2021survey}
N.~Mehrabi {\em et~al.}, ``A survey on bias and fairness in machine learning,'' {\em ACM computing surveys (CSUR)}, 2021.

\bibitem{recht2019imagenet}
B.~Recht, R.~Roelofs, L.~Schmidt, and V.~Shankar, ``Do imagenet classifiers generalize to imagenet?,'' in {\em ICML}, 2019.

\bibitem{ibrahim2022robustness}
M.~Ibrahim, Q.~Garrido, A.~Morcos, and D.~Bouchacourt, ``The robustness limits of sota vision models to natural variation,'' {\em arXiv:2210.13604}, 2022.

\bibitem{feuerriegel2020fair}
S.~Feuerriegel, M.~Dolata, and G.~Schwabe, ``Fair ai: Challenges and opportunities,'' {\em Business \& information systems engineering}, 2020.

\bibitem{hickey2021fairness}
J.~M. Hickey, P.~G. Di~Stefano, and V.~Vasileiou, ``Fairness by explicability and adversarial shap learning,'' in {\em ECML PKDD}, 2021.

\bibitem{pitoura2020social}
E.~Pitoura, ``Social-minded measures of data quality: fairness, diversity, and lack of bias,'' {\em JDIQ}, 2020.

\bibitem{jain2024fairness}
B.~Jain, M.~Huber, and R.~Elmasri, ``Fairness for deep learning predictions using bias parity score based loss function regularization,'' {\em IJAIT}, 2024.

\bibitem{lin2014microsoft}
T.-Y. Lin {\em et~al.}, ``Microsoft coco: Common objects in context,'' in {\em ECCV}, 2014.

\bibitem{gustafson2023facet}
L.~Gustafson {\em et~al.}, ``Facet: Fairness in computer vision evaluation benchmark,'' in {\em ICCV}, 2023.

\bibitem{fedus2022switch}
W.~Fedus, B.~Zoph, and N.~Shazeer, ``Switch transformers: Scaling to trillion parameter models with simple and efficient sparsity,'' {\em JMLR}, 2022.

\bibitem{narayanan2024fairness}
D.~Narayanan {\em et~al.}, ``Fairness perceptions of artificial intelligence: A review and path forward,'' {\em JHCI}, 2024.

\bibitem{ali2024assessing}
S.~Ali {\em et~al.}, ``Assessing generalisability of deep learning-based polyp detection and segmentation methods through a computer vision challenge,'' {\em Scientific Reports}, 2024.

\bibitem{zhang2021understanding}
C.~o. Zhang, ``Understanding deep learning (still) requires rethinking generalization,'' {\em Communications of the ACM}, 2021.

\bibitem{wu2022fair}
X.~Wu, D.~Xu, S.~Yuan, and L.~Zhang, ``Fair data generation and machine learning through generative adversarial networks,'' in {\em Generative Adversarial Learning: Architectures and Applications}, Springer, 2022.

\bibitem{celis2020data}
L.~E. Celis, V.~Keswani, and N.~Vishnoi, ``Data preprocessing to mitigate bias: A maximum entropy based approach,'' in {\em ICML}, 2020.

\bibitem{li2021ditto}
T.~Li, S.~Hu, A.~Beirami, and V.~Smith, ``Ditto: Fair and robust federated learning through personalization,'' in {\em ICML}, 2021.

\bibitem{parraga2023fairness}
O.~Parraga {\em et~al.}, ``Fairness in deep learning: A survey on vision and language research,'' {\em ACM Computing Surveys}, 2023.

\bibitem{terven2023comprehensive}
J.~Terven, D.-M. C{\'o}rdova-Esparza, and J.-A. Romero-Gonz{\'a}lez, ``A comprehensive review of yolo architectures in computer vision: From yolov1 to yolov8 and yolo-nas,'' {\em Machine Learning and Knowledge Extraction}, 2023.

\bibitem{koonce2021mobilenetv3}
B.~Koonce and B.~Koonce, ``Mobilenetv3,'' {\em Convolutional Neural Networks with Swift for Tensorflow: Image Recognition and Dataset Categorization}, 2021.

\bibitem{mehta2021mobilevit}
S.~Mehta and M.~Rastegari, ``Mobilevit: light-weight, general-purpose, and mobile-friendly vision transformer,'' {\em arXiv preprint arXiv:2110.02178}, 2021.

\bibitem{rincy2020ensemble}
T.~N. Rincy and R.~Gupta, ``Ensemble learning techniques and its efficiency in machine learning: A survey,'' in {\em IDEA}, 2020.

\bibitem{du2022glam}
N.~Du {\em et~al.}, ``Glam: Efficient scaling of language models with mixture-of-experts,'' in {\em ICML}, 2022.

\bibitem{riquelme2021scaling}
C.~Riquelme {\em et~al.}, ``Scaling vision with sparse mixture of experts,'' {\em NeurIPS}, 2021.

\bibitem{cui2022synergy}
S.~Cui {\em et~al.}, ``Synergy-of-experts: Collaborate to improve adversarial robustness,'' {\em NeurIPS}, 2022.

\bibitem{aimar2023balanced}
E.~S. Aimar {\em et~al.}, ``Balanced product of calibrated experts for long-tailed recognition,'' in {\em CVPR}, 2023.

\bibitem{vaswani2017attention}
A.~Vaswani {\em et~al.}, ``Attention is all you need,'' {\em NeurIPS}, 2017.

\bibitem{khan2022transformers}
S.~Khan {\em et~al.}, ``Transformers in vision: A survey,'' {\em ACM computing surveys (CSUR)}, 2022.

\bibitem{white2021bananas}
C.~White, W.~Neiswanger, and Y.~Savani, ``Bananas: Bayesian optimization with neural architectures for neural architecture search,'' in {\em AAAI}, 2021.

\bibitem{pelikan2002scalability}
M.~Pelikan, K.~Sastry, and D.~E. Goldberg, ``Scalability of the bayesian optimization algorithm,'' {\em IJAR}, 2002.

\bibitem{chen2022task}
T.~Chen {\em et~al.}, ``Task-specific expert pruning for sparse mixture-of-experts,'' {\em arXiv:2206.00277}, 2022.

\bibitem{huggingface_timm_fastest}
H.~Face, ``Timm top 20 fastest models.'' \url{https://huggingface.co/collections/timm/timm-top-20-fastest-models-655d84afb5da99edaf3a51c3}, 2021.

\bibitem{cai2023efficientvit}
H.~Cai {\em et~al.}, ``Efficientvit: Lightweight multi-scale attention for high-resolution dense prediction,'' in {\em ICCV}, 2023.

\bibitem{chen2022repghost}
C.~Chen {\em et~al.}, ``Repghost: a hardware-efficient ghost module via re-parameterization,'' {\em arXiv:2211.06088}, 2022.

\bibitem{cui2021pp}
C.~Cui {\em et~al.}, ``Pp-lcnet: A lightweight cpu convolutional neural network,'' {\em arXiv preprint arXiv:2109.15099}, 2021.

\bibitem{mehta2022separable}
S.~Mehta and M.~Rastegari, ``Separable self-attention for mobile vision transformers,'' {\em arXiv:2206.02680}, 2022.

\bibitem{sandler2018mobilenetv2}
M.~Sandler {\em et~al.}, ``Mobilenetv2: Inverted residuals and linear bottlenecks,'' in {\em CVPR}, 2018.

\bibitem{he2016deep}
K.~He, X.~Zhang, S.~Ren, and J.~Sun, ``Deep residual learning for image recognition,'' in {\em CVPR}, 2016.

\bibitem{tan2019mnasnet}
M.~Tan {\em et~al.}, ``Mnasnet: Platform-aware neural architecture search for mobile,'' in {\em CVPR}, 2019.

\bibitem{howard2019searching}
A.~Howard {\em et~al.}, ``Searching for mobilenetv3,'' {\em ICCV}, 2019.

\bibitem{han2020model}
K.~Han {\em et~al.}, ``Model rubik’s cube: Twisting resolution, depth and width for tinynets,'' {\em NeurIPS}, 2020.

\bibitem{cai2018proxylessnas}
H.~Cai, L.~Zhu, and S.~Han, ``Proxylessnas: Direct neural architecture search on target task and hardware,'' {\em arXiv:1812.00332}, 2018.

\bibitem{deng2009imagenet}
J.~o. Deng, ``Imagenet: A large-scale hierarchical image database,'' in {\em CVPR}, 2009.

\bibitem{tan2019efficientnet}
M.~Tan and Q.~Le, ``Efficientnet: Rethinking model scaling for convolutional neural networks,'' in {\em ICML}, 2019.

\bibitem{furutanpey2023architectural}
A.~Furutanpey {\em et~al.}, ``Architectural vision for quantum computing in the edge-cloud continuum,'' {\em arXiv preprint arXiv:2305.05238}, 2023.

\end{thebibliography}



\end{document}